Multi-graph Graph Matching for Coronary Artery Semantic Labeling in Invasive Coronary Angiograms

Chen Zhao[1], Zhihui Xu[2], Pukar Baral[3], Michel Esposito[4], and Weihua Zhou[3,5]

[1] Department of Computer Science, Kennesaw State University, Marietta GA, USA

[2] Department of Cardiology, The First Affiliated Hospital of Nanjing Medical University, Nanjing, China

[3] Department of Applied Computing, Michigan Technological University, Houghton, MI, USA

[4] Department of Cardiology, Medical University of South Carolina, Charleston, SC, USA

[5] Center for Biocomputing and Digital Health, Institute of Computing and Cyber-systems, and Health Research Institute, Michigan Technological University, Houghton, MI, USA

Corresponding author

Weihua Zhou, Ph.D.

Department of Applied Computing, Michigan Technological University,

1400 Townsend Dr, Houghton, MI, 49931, USA

Tel: 906-487-2666

E-Mail: whzhou@mtu.edu

**Abstract**. Coronary artery disease (CAD) stands as the leading cause of death worldwide, and invasive coronary angiography (ICA) remains the gold standard for assessing vascular anatomical information. However, deep learning-based methods encounter challenges in generating semantic labels for arterial segments, primarily due to the morphological similarity between arterial branches and varying anatomy of arterial system between different projection view angles and patients. To address this challenge, we model the vascular tree as a graph and propose a multi-graph graph matching (MGM) algorithm for coronary artery semantic labeling. The MGM algorithm assesses the similarity between arterials in multiple vascular tree graphs, considering the cycle consistency between each pair of graphs. As a result, the unannotated arterial segments are appropriately labeled by matching them with annotated segments. Through the incorporation of anatomical graph structure, radiomics features, and semantic mapping, the proposed MGM model achieves an impressive accuracy of 0.9471 for coronary artery semantic labeling using our multi-site dataset with 718 ICAs. With the semantic labeled arteries, an overall accuracy of 0.9155 was achieved for stenosis detection. The proposed MGM presents a novel tool for coronary artery analysis using multiple ICA-derived graphs, offering valuable insights into vascular health and pathology.

**Keywords**: Coronary artery disease · Invasive coronary angiography · Deep learning · Graph matching

# 1. Introduction

Coronary artery disease (CAD) is the leading cause of death worldwide and is associated with 17.8 million deaths annually [1]. In the clinical practice of CAD diagnosis, invasive coronary angiography (ICA) continues to serve as the gold standard [2,3]. This diagnostic technique is crucial for assisting cardiologists in detecting blockages within coronary arteries. With the extracted individual arterial segments, automated identification of these anatomical branches offers valuable insights for generating reports for CAD. This includes quantifying and evaluating stenosis, as well as quantifying regions of interest, such as fractional flow reserve [4]. However, it is important to recognize the inherent limitations associated with the subjective visual assessment of ICAs [5].

Coronary artery segmentation poses challenges at three distinct image levels: surrounding, local and semantic [6]. At the surrounding level, the low contrast between foreground vessels and the background is problematic. This is primarily caused by X-ray power and contrast agent limitations, hindering accurate foreground extraction and leading to incomplete vascular segment delineation. Additionally, difficulties in excluding non-vessel areas arise due to their similar appearance to vascular segments, such as catheter outlines, spines, and ribs, complicating stenosis severity assessment. At the local level, local ambiguity near coronary vessel boundaries poses a challenge. High-frequency detail loss during the 3D to 2D projection in ICA imaging results in smooth grayscale boundaries instead of distinct steps, adversely affecting vessel boundary delineation. This ambiguity is particularly critical in the limited number of pixels near stenosis areas, impacting the accuracy of stenosis severity assessment. Furthermore, due to patient specific anatomy, the variations of projection angles, and contrast dye degradation, recognizing the type of individual coronary arteries remains challenging. According to training guidelines provided by the American College of Cardiology Core Cardiology Training Symposium, it is recommended that trainees spend 4 months and achieve a minimum cumulative number of 100 cases to understand coronary anatomy sufficiently to perform independent diagnostic cardiac catheterization [7]. Thus, manual annotation of coronary arteries is time-consuming and challenging. In conclusion, automated identification of individual arteries is essential for improving efficiency and accuracy in cardiac diagnostics, especially for junior cardiologists and trainees.

The coronary vascular system comprises two major components: the left coronary artery (LCA) and the right coronary artery trees. The clinical significance of the LCA lies in its role as the primary supplier of blood to the left ventricle [8]. The LCA further branches into three main coronary arteries: the left anterior descending (LAD) artery, the left circumflex (LCX) artery, and the left main artery (LMA). The LAD gives rise to diagonal branches (D), while the LCX gives rise to obtuse marginal (OM) branches. However, semantic labeling of coronary arteries faces challenges when relying solely on position and imaging features [6]. The complexity arises from diverse view angles in ICAs and morphological similarities among different artery branches. These challenges underscore the need for innovative approaches in coronary artery semantic labeling.

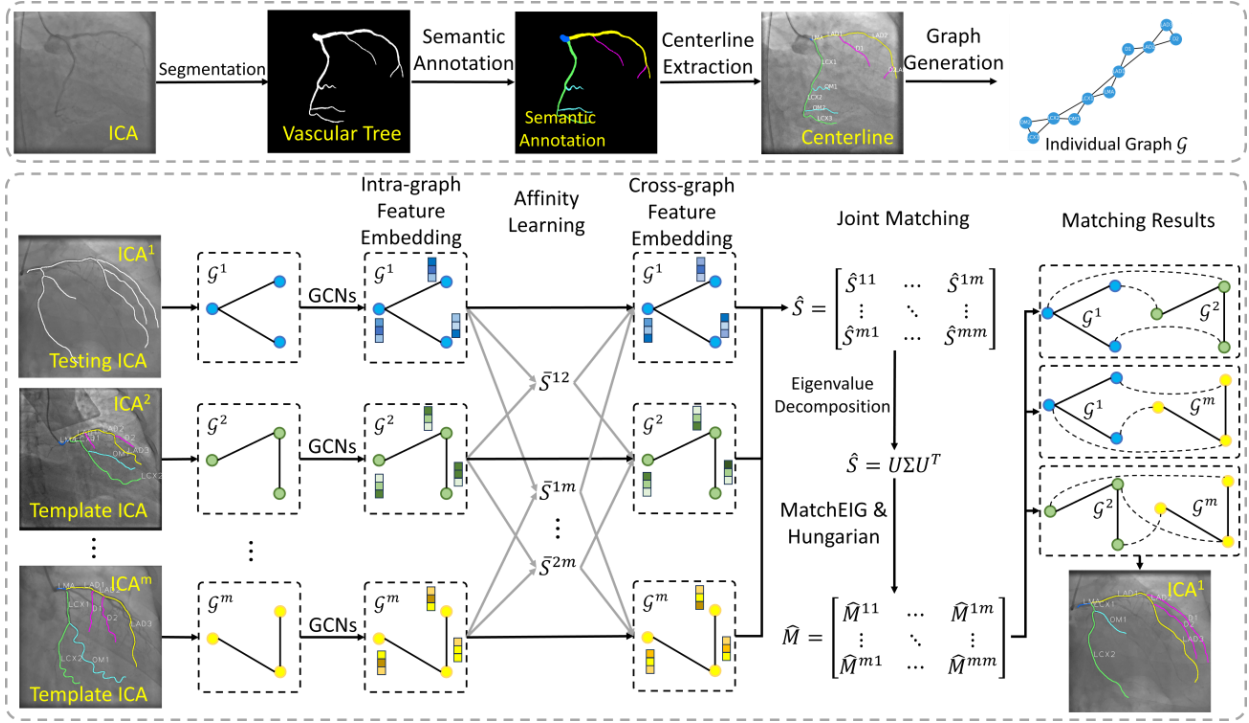

**Figure 1**. Workflow of multi-graph graph matching (MGM) for coronary artery semantic labeling. *Top*: individual graph generation; *Bottom*: MGM among 3 individual graphs derived from a set of ICAs. ICA$^1$ represents the ICA with unlabeled arteries, while ICA$^2$ to ICA$^m$ indicate the labeled template ICAs. MGM considers the cycle consistency between these ICAs and the relationship between arteries from different ICAs will be used to classify the unlabeled arteries in ICA$^1$.

Pixel-to-pixel based approaches face difficulties in identifying categories of coronary arteries due to the morphological and pixel-intensity based feature similarities between different arterial segments. These similarities often lead to ambiguities and misclassifications when relying solely on pixel-level information. In addition, pixel-to-pixel based methods face challenges in identifying small arterial branches. The reduced size and subtle differences in pixel intensity of these branches make them particularly difficult to distinguish from surrounding tissues and larger arteries, leading to frequent misidentifications [5]. In contrast, Zhao et al. employed graph matching to perform coronary artery semantic labeling [9,10], aiming to establish relationships between unlabeled and labeled arterial branches, thus highlighting its potential to address some of the limitations of pixel-based methods. However, their methods have been confined to the comparison of two ICA-generated vascular trees for coronary artery semantic labeling. This approach may not capture the complexity inherent in arterial anatomy.

This paper addresses the critical task of coronary artery semantic labeling using Multi-graph Graph Matching (MGM), which incorporates multiple vascular tree structures obtained from various ICA frames, as shown in Figure 1. The innovation lies in extending graph matching to operate across multiple graphs simultaneously. The arteries in the unlabeled ICA$^1$ are matched with the arteries in the labeled templates ICA$^2$ to ICA$^m$ using the node-to-node correspondence by MGM. The incorporation of cycle consistency further enhances the accuracy of the matching process, offering a sophisticated solution to the challenges posed by the morphological similarity between arterial branches. Experimental results indicated that the proposed MGM-based coronary artery semantic labeling algorithm achieved an average accuracy of 0.9471 on our large scaled multi-centered datasets with 718 ICAs. This paper presents a paradigm shift in coronary artery semantic labeling using multi-ICA joint analysis, opening new avenues for improved understanding and analysis of vascular anatomy.

## 2. Related Work

Modern deep learning based approaches for extracting individual coronary arteries are classified into two categories: semantic segmentation and semantic labeling. Semantic segmentation is a pixel-level labeling task where the goal is to classify each pixel in an image into a specific class/category. For example, in the context of coronary artery segmentation, semantic segmentation would involve labeling each pixel in an angiogram as belonging to one of the coronary arteries (LAD, LCX, and right coronary artery) or background [5,6,11]. However, due to the morphological similarity among different types of arteries, the variability in anatomical structures and varying arterial anatomy under different projection view angles, classifying each pixel without considering topological connectivity between different arterial branches is challenging. The above three approaches [5,6,11] only focused on extracting the main branches. The side branches, such as the D and OM branches, were ignored, even though they are important for understanding the coronary anatomy and making clinical decisions.

Semantic labeling, on the other hand, is a broader task that involves assigning meaningful labels to individual objects or regions of interest within an image. It focuses more on the object-level or region-level labeling rather than pixel-level classification. In addition, the term semantic labeling has been applied to coronary artery identification tasks. By considering the topological information, it is possible to identify the side branches along with the main branches. Yang et al. [12] and Zhang et al. [13] employed a graph neural network and convolutional long-short term memory (LSTM) to classify arterial segments using coronary computed tomography angiograms (CCTA). Wu et al. [14] employed bi-directional tree LSTMs to extract hierarchical features along the arterial tree to perform coronary artery semantic labeling using CCTA. Compared to CCTA, performing coronary artery semantic labeling in ICAs is more challenging due to the nature of high-frequency detail loss during the 3D to 2D projection in ICA imaging, which results in smooth grayscale boundaries instead of distinct steps, adversely affecting vessel boundary delineation. Zhao et al. developed two graph matching based methods using association graph and graph neural network for coronary artery semantic labeling using ICAs [9,10]. However, these two methods only consider the relationship between two ICA generated graphs. In contrast, our multi-graph graph matching method for ICA semantic labeling offers enhanced robustness, accuracy, and global consistency by leveraging information from multiple ICA generated graphs, thereby reducing noise, ambiguities, and ensuring comprehensive insights compared to traditional pairwise graph matching.

## 3. Methodology

### 3.1. Individual Graph Generation

The top of Figure 1 depicts the workflow to generate the individual arterial graph. The application of FP-U-Net++ [15] is utilized for the extraction of the vascular tree, accompanied by the assignment of pixel-level semantic labels by experienced cardiologists. The arterial centerline is obtained through an edge-linking algorithm [16], and then we separate the centerline into different branches to construct the arterial graph, where nodes indicate arterial segments, and edges denote connections. 121 features, including the topology, pixel characteristics, and positional attributes, as detailed in [9], are extracted from each node. Formally, each arterial graph is represented by an undirected acyclic graph $\mathcal{G} = (\mathbb{V}, \mathbb{E}, \mathcal{V})$, where $\mathbb{V} = \{\mathbb{V}_1, \mathbb{V}_2, \cdots, \mathbb{V}_n\}$ represents the node set with $n$ nodes; $\mathbb{E} = \{\mathbb{E}_1, \mathbb{E}_2, \cdots, \mathbb{E}_{n_e}\}$ represents the edge set with $n_e$ edges, containing the connectivity between arterial branches; $\mathcal{V} = \{v_1, v_2, \cdots, v_n\}$ indicates features for nodes, where $v_n \in \mathbb{R}^{121}$.

### 3.2. Multi-graph Graph Matching

The graph matching between two graphs aims to build the node-to-node correspondence between each of the node pairs for each pair of the graphs [17], as denoted as the binary permutation matrix $M^{IJ} \in \mathbb{R}^{n_I \times n_J}$, where $n_I$ and $n_J$ represents the number of nodes in $I$-th and $J$-th graphs in $\mathcal{G}^I$ and $\mathcal{G}^J$. $M_{ij}^{IJ}$ represents the similarity between the two arterial segments, i.e. two nodes $\mathbb{V}_i^I \in \mathcal{G}^I$ and $\mathbb{V}_j^J \in \mathcal{G}^J$. For MGM, we further consider cycle consistency, which denotes a condition where the matching between any two graphs is consistent when passed through any other graphs. Given a set of graphs $\mathbb{G} = \{\mathcal{G}^1, \mathcal{G}^2, \cdots, \mathcal{G}^m\}$ with $m$ individual graphs, the cycle consistency is represented by $M^{IJ} = M^{IK} \cdot M^{KJ}$, s.t. $I, J, K \in \{1, \cdots, m\}$. Note that we use lowercase $i, j$ and $k$ to represent the index of nodes and uppercase $I, J$, and $K$ to represent the index of ICA-generated graphs. By applying the proposed MGM to the graphs generated by one unlabeled ICA and $m - 1$ labeled ICAs, the artery segments are matched, so that semantic labeling is achieved. The proposed MGM model for coronary artery semantic labeling includes 3 modules, and the corresponding computational diagrams are shown in Figure 2.

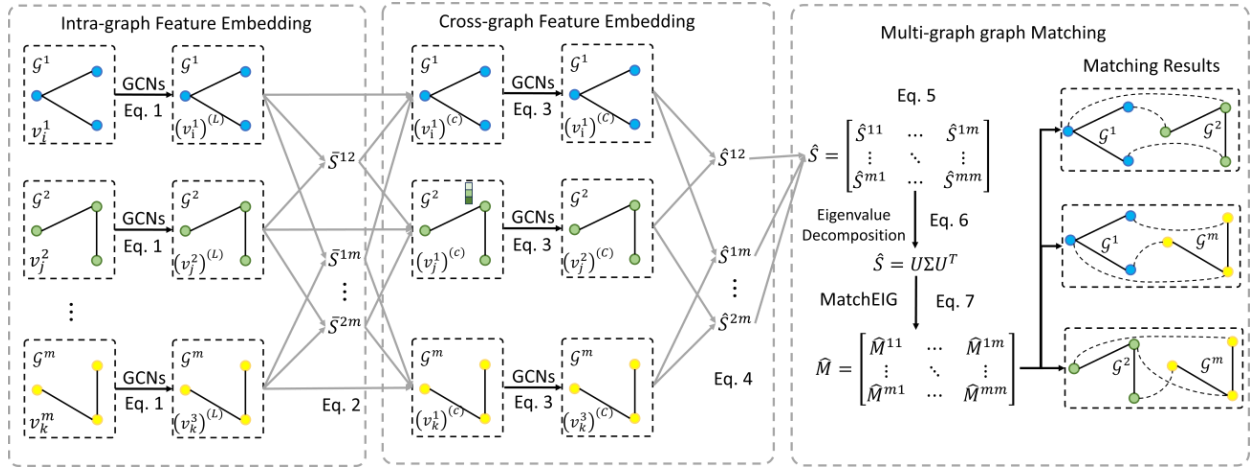

**Figure 2**. Computational diagrams for the proposed MGM. The intra-graph feature embedding is employed to perform feature embedding using GCNs within graphs, while the cross-graph feature embedding is employed to perform feature embedding using GCNs with the affinity generated by the intra-graph feature embedding module. The updated features, after performing intra- and cross-graph feature embedding, are further used to generate the matrix $\hat{S}$ for eigenvalue decomposition and MGM.

1) *Intra-graph Feature Embedding and Affinity Learning*. The feature embedding module aims to capture the essential representation of each arterial segment using graph convolutional networks (GCNs) and multi-layer perceptron (MLP). Given a graph $\mathcal{G}$, the intra-graph feature embedding for node $\mathbb{V}_i$ in $\mathcal{G}$ is denoted in Eq. 1.

$$v_i^{(l+1)} = \phi_v^{(l)}\left(\left[\sum_{j \in E_i} v_j^{(l)}, v_i^{(l)}\right]\right) \qquad (1)$$

where $l \in \{1, \cdots, L\}$ represents the index of GCN layers; $\phi_v^{(l)}$ is the MLP for $l$-th layer; $E_i$ contains the edges connected with node $\mathbb{V}_i$; $\sum_{j \in E_i} \cdot$ indicates the element-wise summation of the features from the adjacent nodes; $[\cdot]$ indicates feature concatenation. Note that we use the upper script with parentheses to denote the index of layers in the corresponding neural network. As a result, Eq. 1 represents the message passing and aggregation for feature embedding using GCN in graph $\mathcal{G}$.

The embedding model described above encodes the structural affinity between two graphs into node-to-node affinity within the embedding space. Given two graphs $\mathcal{G}^I$ and $\mathcal{G}^J$, the intra-graph node similarity is computed by the weighted dot product of the updated feature embedding for each node pair. Since the Sinkhorn operators has already shown effective and efficient performance in permutation prediction [18], we employ a Sinkhorn operator [19] to convert the node similarity matrix into a doubly-stochastic matrix as node affinity, as shown in Eq. 2.

$$\bar{S}_{ij}^{IJ} = Sinkhorn\left(\exp\left(\frac{(v_i^I)^{(L)} \cdot A_{intra} \cdot (v_j^J)^{(L)}}{\sqrt{d_{intra}}}\right)\right) \qquad (2)$$

where $i$ and $j$ are the node indices in $\mathcal{G}^I$ and $\mathcal{G}^J$; $d_{intra}$ is the dimension of embedded features; and $A_{intra} \in \mathbb{R}^{d_{intra} \times d_{intra}}$ contains learnable parameters for the node affinity during intra-graph feature embedding. As a result, the intra-graph node affinity $\bar{S}^{IJ}$ indicates the initial graph matching assignment between each node pair from $\mathcal{G}^I$ and $\mathcal{G}^J$. The exponential form in Eq. 2 is employed to guarantee the non-negative form of the node-to-node affinity, which is required for Sinkhorn algorithm [20].

2) *Cross-graph Feature Embedding in Graph Pairs*. Without interactive cross-graph feature aggregation, the direct node-to-node correspondence lacks reliability [21]. Cross-graph feature embedding refers to the process of characterizing nodes and edges from different graphs, which is beneficial in improving the performance of graph matching [22]. To aggregate node features across graphs, we incorporate intra-graph affinity and cross-graph features using GCNs. Given two graphs $\mathcal{G}^I$ and $\mathcal{G}^J$, the cross-graph feature embedding for node $\mathbb{V}_i$ in $\mathcal{G}^I$ considering all nodes in $\mathcal{G}^J$ is denoted in Eq. 3.

$$(v_i^I)^{(c)} = \psi_v^{(c)}\left(\left[\sum_{j=1}^{n_J} \bar{S}_{ij}^{IJ} \cdot (v_j^J)^{(c)}, (v_i^I)^{(c)}\right]\right) \qquad (3)$$

where $\psi_v^{(c)}$ is the MLP used in cross-graph embedding module for $c$-th layer and $c \in \{0,\cdots,C\}$. If $c = 0$, then $(v_i^I)^{(c)} = (v_i^I)^{(L)}$. We adopted the cross-graph feature embedding in Eq. 3 for $C$ times to extract cross-graph hierarchical features and perform non-linear transformations. Symmetrically, for $\mathbb{V}_j$ in $\mathcal{G}^J$, the updated cross-graph feature embedding is calculated as $(v_j^J)^{(c)} = \psi_v^{(c)}\left(\left[\sum_{j=1}^{n_J} \bar{S}_{ij}^{IJ} \cdot (v_i^I)^{(c)}, (v_j^J)^{(c)}\right]\right)$. Finally, the affinity matrix is updated according to the updated cross-graph node feature embedding, as shown in Eq. 4.

$$\hat{S}_{ij}^{IJ} = Sinkhorn\left(\exp\left(\frac{(v_i^I)^{(C)} \cdot A_{cross} \cdot (v_j^J)^{(C)}}{\sqrt{d_{cross}}}\right)\right) \qquad (4)$$

where $d_{cross}$ is the dimension of embedded features during cross-graph feature embedding. And $A_{cross} \in \mathbb{R}^{d_{cross} \times d_{cross}}$ contains learnable parameters for the node affinity during intra-graph feature embedding.

3) *Multi-graph graph matching*. We first perform the feature embedding for each graph in $\mathbb{G}$ according to Eq. 1. For each pair of graphs in $\mathbb{G}$, denoted as $\mathcal{G}^I$ and $\mathcal{G}^J$, we calculate the joint node affinity matrix $\hat{S}_{ij}^{IJ}$ according to Eqs. 2 to 4 using cross-graph feature embedding. In MGM, we hypothesize that each graph contains the same number of nodes, i.e. $n_1 = n_2 = \cdots = n_m = n$. Thus, $\hat{S}^{IJ} \in \mathbb{R}^{n \times n}$. For $m$ graphs, the graph-matching results are denoted as a joint matching matrix $\hat{S} \in \mathbb{R}^{nm \times nm}$, as defined in Eq. 5.

$$\hat{S} = \begin{pmatrix} \hat{S}^{11} & \cdots & \hat{S}^{1m} \\ \vdots & \ddots & \vdots \\ \hat{S}^{m1} & \cdots & \hat{S}^{mm} \end{pmatrix} \tag{5}$$

where the diagonal-sub matrices $\hat{S}^{II}$ are identical matrices s.t. $I \in [1,\cdots,m]$. During the MGM computation, we only measure the upper diagonal elements in $\hat{S}$ and use transposed matrix as the lower diagonal elements, i.e. $\hat{S}^{IJ} = (\hat{S}^{JI})^T$. As a result, $\hat{S}$ is a symmetric matrix.

Given $m$ graphs, the hard assignment of the graph matching, i.e. ground truth is denoted as $M = [M^{11}, M^{12}, \cdots, M^{1m}]^T \in \mathbb{R}^{mn \times n}$, where each submatrix squared matrix in $M^{1K}, s.t. K \in [1, \cdots, m]$ with $n^2$ elements is a binary permutation matrix. The goal of MGM is to maximize the matching score, as $S = M^T \hat{S} M$. Since $\hat{S}$ is a symmetric matrix, according to Rayleigh's ratio theorem, the $M$ that will maximize $S$ is the principal eigenvector of $\hat{S}$ [23]. Thus, the MGM solution can be obtained by computing the maximum eigenvector of $\hat{S}$ [24]. Mathematically, the eigenvector decomposition of $\hat{S}$ is denoted in Eq. 6.

$$\hat{S} = U \Sigma U^T \tag{6}$$

where $U = [\hat{U}^{1K}, \hat{U}^{2K}, \cdots, \hat{U}^{mK}]^T \in \mathbb{R}^{mn \times n}$ is the $n$ corresponding eigenvectors and $K \in [1, \cdots m]$; the diagonal matrix $\Sigma \in \mathbb{R}^{n \times n}$ contains top-$n$ eigenvalues. Splitting the eigenvectors $U$ to $\hat{U}^{iK} \in \mathbb{R}^{n \times n}$, we apply the MatchEIG algorithm [25] to generate the refined permutation matrix while considering the cycle consistency during the MGM. The predicted hard-assignment matrix between graphs $\mathcal{G}^I$ and $\mathcal{G}^J$ is denoted in Eq. 7.

$$\widehat{M}^{IJ} = Hungarian\left(\left(\hat{U}^{IJ}\right) \cdot \left(\hat{U}^{IJ}\right)^T\right) \tag{7}$$

where the Hungarian algorithm [26] is commonly employed as a post-processing step to transform the doubly-stochastic matrix into the permutation matrix $\widehat{M}^{IJ}$. To guarantee the cycle constancy, an arbitrary $K$ is randomly selected [27].

### 3.3. Computational workflow and model training

According to Section 3.2, the proposed MGM contains 3 sub modules, and the intra-graph and cross-graph modules are iteratively invoked. The overall computation workflow of MGM is shown in Figure 2.

*Optimization* To train the proposed MGM, a set of graphs $\mathbb{G}$ with $m$ graphs is prepared. According to the annotated ICAs, the node correspondences between each arterial segment are obtained, resulting in the ground truth permutation matrix $M^{IJ}$ for each pair of graphs $\mathcal{G}^I$ and $\mathcal{G}^J$. If the arterial nodes $\mathbb{V}_i^I \in \mathcal{G}^I$ and $\mathbb{V}_j^J \in \mathcal{G}^J$ have identical labels, then $M_{ij}^{IJ} = 1$. To ensure consistency and accuracy, we select individual graphs exclusively from ICAs captured with the same view angle and the same number of arterial segments. The cross-entropy loss between the predicted permutation matrix $\widehat{M}^{IJ}$ and the ground truth $M^{IJ}$ for each pair of graphs in the MGM setting is employed as the objective function, as defined in Eq. 8.

$$L = -\sum_{I}^{m} \sum_{J>I}^{m} \sum_{i=1}^{n} \sum_{j=1}^{n} (1 - M_{ij}^{IJ}) \log(1 - \widehat{M}^{IJ}) + (\widehat{M}^{IJ}) \log(\widehat{M}^{IJ}) \tag{8}$$

where the second sub-index $J > I$ indicates we only calculate the graph matching cross entropy loss belonging to the upper triangular elements. The overall MGM training algorithm is shown in Algorithm 1.

**Algorithm 1**. The training process of an epoch for MGM for coronary artery semantic labeling

---

**Inputs**: $D_{tr}$ with $N_{tr}$ ICAs; $D_{tp}$ with $N_{tp}$ template ICAs; $m$: the number of MGM graphs;

$L$: number of intra-graph embedding layers;

$C$: number of cross-graph embedding layers;

**Outputs**: updated MGM

For $I = 1$ to $N_{tr}$:

1. For $J_1, J_2, \cdots, J_{m-1} = Combination(N_{tp}, m - 1)$:
2.    If $\mathcal{G}^I_{tr}, \mathcal{G}^{J_1}_{tp}, \cdots, \mathcal{G}^{J_{m-1}}_{tp}$ have same view angles and number of nodes:
3.       Extract intra-graph features for $\mathcal{G}^I_{te}, \mathcal{G}^{J_1}_{tp}, \cdots, \mathcal{G}^{J_{m-1}}_{tp}$ for $L$ layers;
4.       Extract cross-graph features between each pair from $\mathcal{G}^I_{te}, \mathcal{G}^{J_1}_{tp}, \cdots, \mathcal{G}^{J_{m-1}}_{tp}$ for $C$ layers;
5.       Build joint matching matrix $\hat{S}$ using Eqs. 2 to 5;
6.       Perform eigenvector decomposition and use Hungarian algorithm to calculate $\hat{M}^{IJ}$;
7.       Calculate loss defined in Eq. 8 according to $\hat{M}^{IJ}$ and $M^{IJ}$;
8.    End if
9. End for

---

### 3.4. Testing

We first separate the entire dataset into three subsets, as $D_{tr}$, $D_{te}$ and $D_{tp}$ for the training set, the testing set and the template set with $N_{tr}$, $N_{te}$ and $N_{tp}$ ICAs, respectively. $D_{tp}$ contains a set of representative ICAs selected through stratified sampling based on the first-view and second-view angles of the ICAs. For each patient, a frame typically used in clinical practice for anatomical structure analysis was selected from the video for semantic labeling. The coronary angiogram frame chosen corresponds to the end of the diastolic phase and is automatically identified using our algorithm designed to detect end-diastolic and end-systolic frames in invasive coronary angiography videos [28]. This frame selection ensures optimal visibility and alignment of coronary structures, allowing for accurate segment labeling across similar coronary trees.

We employ MGM to simulate the learning process, where cardiologists learn coronary artery anatomy by comparing a test case to multiple reference cases in $D_{tp}$. Given the intricate anatomy, clinical decisions hinge on multiple ICAs. During the testing, the tested ICA from $D_{te}$ is denoted as $\mathcal{G}^{te}$ and the other $m - 1$ ICAs are selected from $D_{tp}$ and used to perform MGM; then a majority voting strategy is employed to assign semantic labels to matched arterial branches in $D_{tp}$.

However, using the combination function to find all possible graph matching pairs would generate a great number of matched pairs, which results in slow prediction procedure during the testing, hinder the clinical applicability of the MGM in practice, as each tested ICA in $D_{te}$ is compared with $Combination(N_{tp}, m - 1)$ ICAs in $D_{tp}$.

In the coronary artery system, arterial pathways strictly adhere to cardiovascular anatomy. For instance, initial segments of LCX and LAD arteries connect to LMA, side D branches link to LAD, and side OM branches connect to LCX. Deviations from this anatomical structure in predicted arterial labels incur a weighted penalty on graph-matching outcomes. We developed a lookup table storing physical connections between arterial branches. A prediction is deemed confident if its connected arteries match those in the lookup table. Conversely, discrepancies trigger a penalty proportional to the percentage of erroneous connections. For example, if the predicted LAD is connected to LMA and LCX, we accept this prediction. Otherwise, if the predicted OM is connected to LMA, we reject this graph matching result. Only the graph matching results without structural loss are considered during majority voting. The testing algorithm is shown in Algorithm 2.

**Algorithm 2**. The testing process of MGM for coronary artery semantic labeling

---

**Inputs**: $D_{te}$ with $N_{te}$ ICAs; $D_{tp}$ with $N_{tp}$ template ICAs; $m$: the number of MGM graphs

**Outputs**: labels for each arterial segment of $N_{te}$ ICAs in $D_{te}$

For $I = 1$ to $N_{te}$:

1. For $J_1, J_2, \cdots, J_{m-1} = Combination(N_{tp}, m-1)$:
2.     If $\mathcal{G}_{te}^I, \mathcal{G}_{tp}^{J_1}, \cdots, \mathcal{G}_{tp}^{J_{m-1}}$ have same view angles and number of nodes:
3.         Extract intra-graph features for $\mathcal{G}_{te}^I, \mathcal{G}_{tp}^{J_1}, \cdots, \mathcal{G}_{tp}^{J_{m-1}}$;
4.         Extract cross-graph features between each pair from $\mathcal{G}_{te}^I, \mathcal{G}_{tp}^{J_1}, \cdots, \mathcal{G}_{tp}^{J_{m-1}}$;
5.         Build joint matching matrix $\hat{S}$ using Eqs. 2 to 5;
6.         Perform eigenvector decomposition and graph matching using Eqs. 6 and 7;
7.         Save $\hat{M}^{I,J_1}, \cdots, \hat{M}^{I,J_{m-1}}$ for majority voting if no structural loss is presented;
8.     End if
9. End for
10. Assign labels to $\mathcal{G}_{te}^I$ according to the majority voting among the saved set of $\hat{M}^{I,J_*}$.

End for

---

### 3.5. Stenosis detection

As illustrated in the introduction, the aim of semantic labeling is to identify the lesion in a specific coronary arterial segment to guide percutaneous coronary intervention and coronary artery bypass grafting for CAD diagnosis and treatment. We developed a comprehensive algorithm for coronary artery stenosis detection, which begins with extracting the arterial centerline from the segmented arterial contours. A morphological algorithm, specifically an iterative erosion operation, is employed to extract the centerline, preserving the vascular tree's connectivity and topology. The diameters of the arterial segments are calculated using the distance transform algorithm.

For stenosis detection, the algorithm utilizes the sequence of diameters along the points in the centerline of an arterial segment. The stenosis detection task is then transformed into a search for the local minimum values in the diameter sequence. We adopt the backward difference method to identify diameter points with local minimum and maximum values according to the discrete diameter sequence. For each point, the second derivative is calculated: points with a second derivative of -2 are defined as local minimum points, and those with a second derivative of 2 are defined as local maximum points. All local minimum points form a candidate list, and the point with the smallest diameter in this list is denoted as the minimum diameter of the arterial segment, denoted as $d_{min}$. Similarly, all local maximum points form another candidate list, and the point with the largest diameter in this list is denoted as the maximum diameter of the arterial segment, denoted as $d_{max}$. The stenotic level is defined as $\frac{d_{min}}{d_{max}} \times 100\%$. Following the standards recommended by the Society of Cardiovascular Computed Tomography [29], stenosis grades are classified into minimal, mild, moderate, and severe, corresponding to 1%–24%, 25%–49%, 50%–69%, and 70%–100%, respectively.

### 3.6. Evaluation

The task of semantic labeling is redefined as a multi-class classification problem among various arterial segments. In this context, the model's performance is assessed using weighted metrics: accuracy (ACC), precision (PREC), recall (REC), and the F1-score (F1). During model training, the LAD and LCX branches are divided into sub-segments since they are separated by side branches, such as D and OM segments. However, these sub-segments are regrouped into their original categories during the evaluation phase. The formulas for weighted ACC, PREC, REC, and F1 are shown in Eqs. 9 to 12.

$$ACC = \frac{1}{N} \sum_{c=1}^{C} \frac{TP_c + TN_c}{TP_c + TN_c + FP_c + FN_c} \times N_c \tag{9}$$

$$PREC = \frac{1}{N} \sum_{c=1}^{C} \frac{TP_c}{TP_c + FP_c} \times N_c \tag{10}$$

$$REC = \frac{1}{N} \sum_{c=1}^{C} \frac{TN_c}{TN_c + FN_c} \times N_c \tag{11}$$

$$F1 = \frac{1}{N} \sum_{c=1}^{C} \frac{TP_c}{TP_c + \frac{1}{2}(FP_c + FN_c)} \times N_c \tag{12}$$

In these equations, $TP_c$, $TN_c$, $FP_c$, and $FN_c$ stand for the true positive, true negative, false positive, and false negative arterial segments, respectively. The variable $c$ denotes the class index of the arterial segments, $C$ is the total number of classes, $N_c$ represents the number of arterial segments in class $c$ and $N$ is the total number of arterial segments.

To evaluate the stenosis detection performance, we measure the accuracy of the correctly identified coronary artery segments with stenotic lesions, as shown in Eq. 13.

$$ACC_s = \frac{1}{N} \sum_{c=1}^{C} \frac{TP_c^s}{TP_c^s + FN_c^s} \times N_c \tag{13}$$

where $TP_c^s$ indicates the number of correctly matched stenotic arterial segment by MGM for $c$-th categorial arterial segments, while $FN_c^s$ represents the number of mismatched stenotic arterial segment by MGM.

## 4. Experimental Results

### 4.1. Dataset and Enrolled Subjects

This study includes 263 ICAs from the First Affiliated Hospital, Nanjing and 455 from Medical University of South Carolina. Both centers obtained ethical approval to retrieve and anonymize clinical data. Experienced cardiologists annotated coronary artery semantic labels, including left main artery (LMA), LAD, LCX, OM, and D branches, if present. All images were resized to 512 × 512. For each patient, a frame that was used for anatomical structure analysis in clinical practice was selected from the view video for semantic labeling. The coronary angiogram frame is chosen at the end of the diastolic phase from the coronary angiography video and the frame index is automatically determined using our developed algorithm for end-diastolic and end-systolic cardiac frames from ICA videos [28]. This specific frame selection is based on the optimal visibility and alignment of coronary structures, ensuring reliable segment labeling between similar coronary trees. For instance, both flow fraction reserve calculated from the 3D arterial model [30] and 3D quantitative coronary angiography [31] require end-systolic image frames in ICA videos.

In detail, we selected the ICA images from 6 view angles, including left anterior oblique cranial, left anterior oblique caudal, right anterior oblique cranial, right anterior oblique caudal, and anterior-posterior cranial, anterior-posterior caudal, as demonstrated in Table 1. During model training, we only selected ICAs from the same view angle to generate pairs for graph matching. During testing, the tested ICA has the identical view angle as the template ICA. As a result, reliable artery-to-artery correspondence is guaranteed. Table 1 displays image counts for each view angle.

**Table 1**. View angles and number of enrolled subjects. CRA, cranial; CAU, caudal; LAO, left anterior oblique; RAO, right anterior oblique; AP: anterior-posterior.

| Site | View Angle | LAO | RAO | AP | TOTAL |
|---|---|---|---|---|---|
| S1 | CRA | 42 | 19 | 18 | 79 |
|    | CAU | 18 | 116 | 56 | 190 |
| S2 | CRA | 44 | 16 | 123 | 183 |
|    | CAU | 20 | 220 | 26 | 266 |
| TOTAL | | 124 | 371 | 223 | 718 |

### 4.2. Implementation Details

All experiments were conducted with an NVIDIA RTX 4090 GPU on a Core I9 14900K. The model was implemented using PyTorch 1.10 built on Python 3.8. We conducted the grid search to determine the optimal hyperparameters. The hyperparameters were tuned to the test set during the cross-validation process for each hyperparameter setting. The grid search settings are shown in Table 2. More specifically, we first set $L = 3$ and tuned $C$ respectively. Then, we fixed $C = 3$ and tuned $L$. After finding the optimal $C$ and $L$, we further tuned $d_{intra}$ and $d_{cross}$. As a results, the number of GCN layers was set as $L = 3$ in Eq. 1 and the number of cross-graph feature embedding layers was set as $C = 3$ in Eq. 3. The dimension of embedded features in Eqs. 2 and 4 was set as $d_{intra} = d_{cross} = 256$. An Adam optimizer with an initial learning rate of $1e-5$ was employed to train the model. 10% ($N_{tp} = 71$) of representative ICAs determined by stratified sampling were selected as the template set and the remaining 90% of ICAs were separated to $D_{tr}$ and $D_{te}$ to perform a five-fold cross-validation to evaluate the model performance.

**Table 2**. Hyperparameter settings in the grid search

| Hyperparameter | Searching space | Description |
|---|---|---|
| $L$ | [1,2,3,4] | Number of GCN layers in intra-graph feature embedding. |
| $C$ | [1,2,3,4] | Number of GCN layers in cross-graph feature embedding. |
| $d_{intra}$ | [128, 256, 512] | The dimension of intra-graph feature embedding. |
| $d_{cross}$ | [128, 256, 512] | The dimension of cross-graph feature embedding. |

### 4.3. MGM for Coronary Artery Semantic Labeling

Considering the computational resource limitations, we set the number of graphs in $\mathbb{G}$ as $m = \{3, 4, 5, 6\}$. The average performance of the five folds is reported in Table 3. As per Table 3, the proposed MGM demonstrated impressive coronary artery semantic labeling performance with an ACC of 0.9471 when comparing three graphs. However, with an increase to four compared graphs, the performance slightly declined to the ACCs of 0.9304 and 0.9352 when $m = 4$ and $m = 5$, respectively. This decline stems from the limited dataset size, posing challenges for ensuring cycle consistency and impacting model performance. Experiments for $m = 6$ showed further deterioration, with an ACC and 0.8832.

With the increase of $m$, it is observed that the performance of the MGM model gradually degrades. This degradation is attributed to the diminishing efficacy of cycle-consistency in ensuring accurate matching as the number of graphs increases. Cycle-consistency, a key component in the multi-graph graph matching algorithm, becomes less reliable with higher values of $m$, leading to decreased performance metrics across various artery types. Moreover, the limited number of available data exacerbates this issue. Note that the $m$ graphs for graph matching should have the same view angles and number of nodes. Specifically, for $m \geq 7$, 44 testing ICAs, equivalent to 34.7% of the testing set, fail to find the required additional $m - 1$ ICAs within the template set to evaluate performance accurately. This significant limitation in the dataset means that beyond $m = 6$, there is an insufficient number of ICAs to sustain the testing process effectively. Consequently, we decided not to increase $m$ further for multi-graph graph matching evaluations, as the results would not be representative due to the data scarcity and the inherent limitations of maintaining cycle-consistency across a larger number of graphs.

**Table 3**. Achieved performance for coronary artery semantic labeling using MGM

| $m$ | Artery | LMA | LAD | LCX | D | OM | Avg |
|---|---|---|---|---|---|---|---|
| 3 | ACC | 0.9962±0.0047 | 0.9618±0.0094 | 0.9396±0.0244 | 0.9459±0.0126 | 0.9096±0.0357 | 0.9471±0.0175 |
|   | PREC | 0.9962±0.0047 | 0.9605±0.0096 | 0.9409±0.0243 | 0.9440±0.0129 | 0.9115±0.0354 | 0.9506±0.0163 |
|   | REC | 0.9962±0.0047 | 0.9618±0.0094 | 0.9396±0.0244 | 0.9459±0.0126 | 0.9096±0.0357 | 0.9506±0.0163 |
|   | F1 | 0.9962±0.0047 | 0.9612±0.0094 | 0.9402±0.0244 | 0.9449±0.0127 | 0.9105±0.0356 | 0.9506±0.0163 |
| 4 | ACC | 0.9941±0.0079 | 0.9414±0.0141 | 0.9299±0.0183 | 0.9172±0.0194 | 0.8930±0.0316 | 0.9304±0.0122 |
|   | PREC | 0.9941±0.0079 | 0.9426±0.0116 | 0.9293±0.0197 | 0.9135±0.0215 | 0.8958±0.0286 | 0.9351±0.0123 |
|   | REC | 0.9941±0.0079 | 0.9414±0.0141 | 0.9299±0.0183 | 0.9172±0.0194 | 0.8930±0.0316 | 0.9351±0.0122 |
|   | F1 | 0.9941±0.0079 | 0.9420±0.0127 | 0.9296±0.0190 | 0.9153±0.0203 | 0.8944±0.0301 | 0.9351±0.0123 |
| 5 | ACC | 0.9887±0.0176 | 0.9476±0.0102 | 0.9356±0.0173 | 0.9208±0.0153 | 0.9015±0.0261 | 0.9352±0.0145 |
|   | PREC | 0.9887±0.0176 | 0.9511±0.0076 | 0.9321±0.0215 | 0.9260±0.0115 | 0.8965±0.0320 | 0.9389±0.0150 |
|   | REC | 0.9887±0.0176 | 0.9476±0.0102 | 0.9356±0.0173 | 0.9208±0.0153 | 0.9015±0.0261 | 0.9389±0.0151 |
|   | F1 | 0.9887±0.0176 | 0.9493±0.0086 | 0.9339±0.0193 | 0.9234±0.0129 | 0.8990±0.0290 | 0.9389±0.0150 |
| 6 | ACC | 0.9500±0.0500 | 0.9067±0.0490 | 0.8933±0.0105 | 0.8635±0.0694 | 0.8141±0.0310 | 0.8832±0.0393 |
|   | PREC | 0.9500±0.0500 | 0.9156±0.0479 | 0.8770±0.0306 | 0.8623±0.0539 | 0.8263±0.0189 | 0.8862±0.0402 |
|   | REC | 0.9500±0.0500 | 0.9067±0.0490 | 0.8933±0.0105 | 0.8635±0.0694 | 0.8141±0.0310 | 0.8855±0.0420 |
|   | F1 | 0.9500±0.0500 | 0.9111±0.0484 | 0.8850±0.0207 | 0.8628±0.0616 | 0.8201±0.0250 | 0.8858±0.0412 |

We evaluated the performance across different view angles, as shown in Table 4. The results indicate that our proposed model generally performs well on most ICA images from various view angles, except for the RAO CRA view angle. It is important to note that during the generation of graph matching pairs, only ICA images from the specific sampled view angle are used for training, while templates from the same view angle are utilized for testing and prediction. The lower performance observed in RAO CRA can be explained by the limited number of training samples available for this view angle, as shown in Table 1. This lack of training data restricts the number of eligible graph matching pairs during training. Furthermore, during testing, the limited number of templates under the LAO CAU view angle leads to biased predictions in graph matching.

**Table 4**. The achieved ACC, PREC, REC and F1 of the proposed MGM using ICAs under different view angles. The $N_c$ indicates the number of ICAs in the testing set among the 5-fold cross-validation under different view angles.

| First View Angle | Second View Angle | ACC | PREC | REC | F1 | $N_c$ |
|---|---|---|---|---|---|---|
| AP | CAU | 0.9690±0.0333 | 0.9715±0.0305 | 0.9715±0.0305 | 0.9715±0.0305 | 39 |
| AP | CRA | 0.9445±0.0207 | 0.9474±0.0210 | 0.9476±0.0207 | 0.9475±0.0209 | 103 |
| LAO | CAU | 0.9643±0.0619 | 0.9625±0.0650 | 0.9625±0.0650 | 0.9625±0.0650 | 27 |
| LAO | CRA | 0.9493±0.0088 | 0.9535±0.0080 | 0.9535±0.0080 | 0.9535±0.0080 | 54 |
| RAO | CAU | 0.9530±0.0161 | 0.9561±0.0152 | 0.9560±0.0152 | 0.9560±0.0152 | 277 |
| RAO | CRA | 0.8594±0.0974 | 0.8752±0.0860 | 0.8752±0.0860 | 0.8752±0.0860 | 21 |

We further visualized the representative graph matching results under different view angles, as illustrated in Figure. 3.

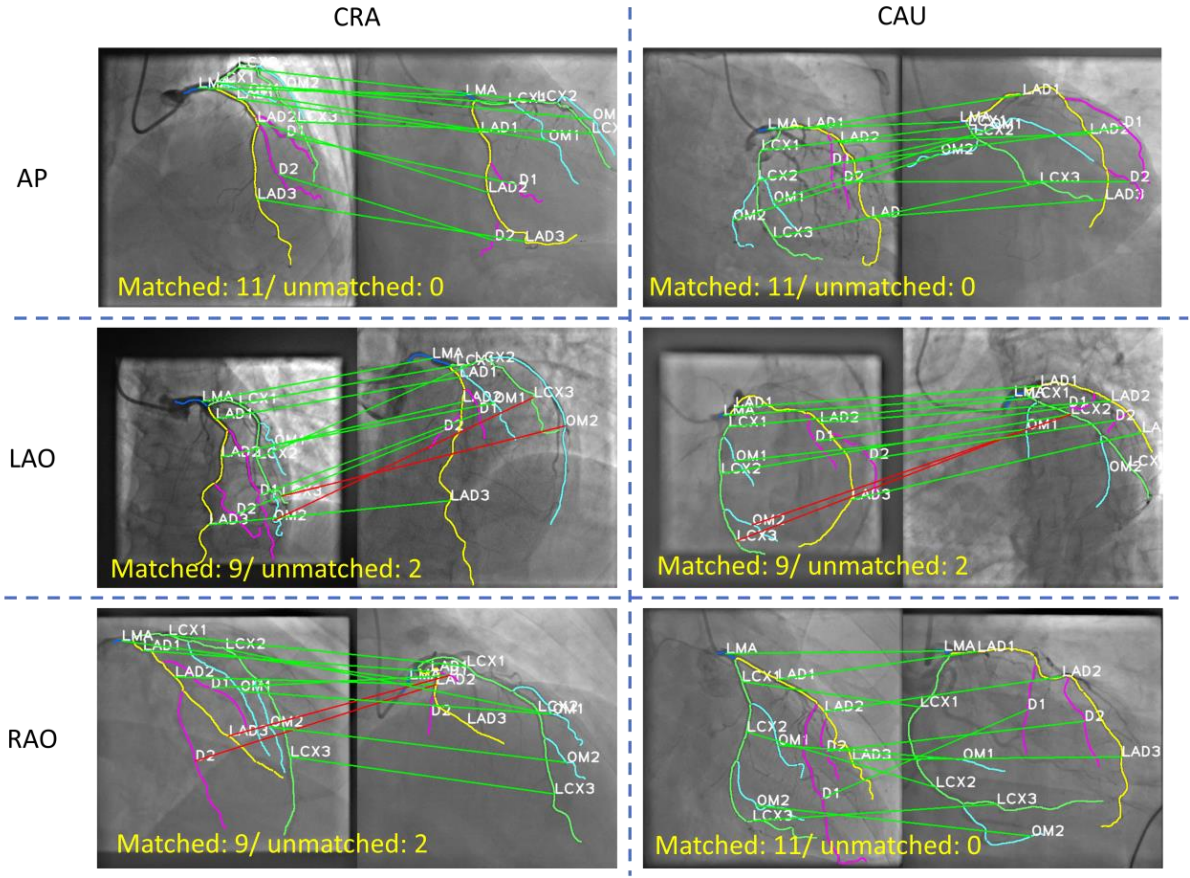

**Figure 3**. Visualization of representative graph matching results. The green line indicates a correct match, while the red line represents an error.

### 4.4. Comparison with Competing Methods

Our MGM approach is compared with four deep learning-based methods for coronary artery semantic labeling:

- Association Graph-based Graph Matching Network (AGMN) [9], utilizing the association graph for two-graph matching;
- Edge Attention Graph Matching Network (EAGMN) [10], an extension of AGMN with added edge attention;
- Neural Graph Matching (NGM) [27], using the association graph-induced affinity matrix for two-graph matching;
- Bidirectional Tree-LSTM (BiTL) [32], employing tree-structured bidirectional LSTM for hierarchical feature extraction along with the vascular tree.

Table 5 illustrates the ACC, PREC, REC and F1 comparison among different methods.

**Table 5**. Achieved performance for coronary artery semantic labeling using MGM. The bold texts indicate the best performance achieved in the corresponding metric and types of arterial segments.

| Model | Metric | LMA | LAD | LCX | D | OM | Avg |
|---|---|---|---|---|---|---|---|
| BiTL  |     | 0.6000±0.4899 | 0.9385±0.0082 | 0.6770±0.2134 | 0.7395±0.3706 | 0.5324±0.3152 | 0.7291±0.0728 |
| AGMN  |     | 0.9907±0.0031 | 0.8730±0.0440 | 0.8646±0.0274 | 0.8320±0.0391 | 0.8080±0.0411 | 0.8639±0.0182 |
| EAGMN | ACC | 0.9942±0.0064 | 0.8931±0.0295 | 0.8843±0.0393 | 0.8518±0.0373 | 0.7968±0.0431 | 0.8767±0.0188 |
| NGM   |     | 0.9885±0.0122 | 0.9223±0.0341 | 0.8967±0.0324 | 0.9007±0.0466 | 0.8408±0.0480 | 0.9039±0.0354 |
| MGM   |     | **0.9962±0.0047** | **0.9618±0.0094** | **0.9396±0.0244** | **0.9459±0.0126** | **0.9096±0.0357** | **0.9471±0.0175** |
| BiTL  |      | 0.6000±0.4899 | 0.8911±0.0661 | 0.6244±0.0906 | 0.6047±0.3027 | 0.5739±0.2876 | 0.6588±0.1504 |
| AGMN  |      | 0.9915±0.0045 | 0.8779±0.0212 | 0.8689±0.0183 | 0.8219±0.0225 | 0.8040±0.0257 | 0.8728±0.0161 |
| EAGMN | PREC | 0.9923±0.0094 | 0.8973±0.0221 | 0.8759±0.0179 | 0.8447±0.0300 | 0.8105±0.0276 | 0.8841±0.0166 |
| NGM   |      | 0.9893±0.0120 | 0.9207±0.0332 | 0.8962±0.0327 | 0.8912±0.0488 | 0.8530±0.0445 | 0.9101±0.0337 |
| MGM   |      | **0.9962±0.0047** | **0.9612±0.0094** | **0.9402±0.0244** | **0.9449±0.0127** | **0.9105±0.0356** | **0.9506±0.0163** |
| BiTL  |     | 0.6000±0.4899 | 0.8574±0.1210 | 0.6780±0.1679 | 0.5115±0.2560 | 0.6991±0.3759 | 0.6692±0.1795 |
| AGMN  |     | 0.9923±0.0097 | 0.8847±0.0188 | 0.8739±0.0224 | 0.8158±0.0501 | 0.8015±0.0301 | 0.8736±0.0163 |
| EAGMN | REC | 0.9904±0.0125 | 0.9017±0.0192 | 0.8690±0.0208 | 0.8386±0.0332 | 0.8284±0.0520 | 0.8856±0.0176 |
| NGM   |     | 0.9901±0.0119 | 0.9192±0.0329 | 0.8957±0.0330 | 0.8820±0.0514 | 0.8657±0.0418 | 0.9105±0.0337 |
| MGM   |     | **0.9962±0.0047** | **0.9605±0.0096** | **0.9409±0.0243** | **0.9440±0.0129** | **0.9115±0.0354** | **0.9506±0.0163** |
| BiTL  |    | 0.6000±0.4899 | 0.9385±0.0082 | 0.6770±0.2134 | 0.7395±0.3706 | 0.5324±0.3152 | 0.6975±0.1160 |
| AGMN  |    | 0.9907±0.0031 | 0.8730±0.0440 | 0.8646±0.0274 | 0.8320±0.0391 | 0.8080±0.0411 | 0.8737±0.0150 |
| EAGMN | F1 | 0.9942±0.0064 | 0.8931±0.0295 | 0.8843±0.0393 | 0.8518±0.0373 | 0.7968±0.0431 | 0.8840±0.0170 |
| NGM   |    | 0.9885±0.0122 | 0.9223±0.0341 | 0.8967±0.0324 | 0.9007±0.0466 | 0.8408±0.0480 | 0.9098±0.0337 |
| MGM   |    | **0.9962±0.0047** | **0.9618±0.0094** | **0.9396±0.0244** | **0.9459±0.0126** | **0.9096±0.0357** | **0.9506±0.0163** |

Table 5 illustrates that our proposed MGM attained the highest ACC of 94.71%. While BiTL, designed for 3D CCTA datasets, proved unsuitable for ICAs, emphasizing the challenge of labeling coronary arteries from 2D images. The proposed MGM model consistently outperformed all other models, demonstrating the highest average scores across all metrics and artery types. This superior performance indicates that MGM is the most robust and reliable model, capable of providing accurate and consistent results. In contrast, the BiTL model showed significant variability, with high performance on the LAD artery but poor results on others, particularly OM. Both AGMN and EAGMN models displayed consistent performance, with EAGMN showing slight improvements over AGMN. The NGM model also performed well, particularly excelling in LAD. However, MGM's advanced multi-graph matching algorithm, which leverages information from multiple graphs, significantly enhances its matching accuracy, making it the most effective model for coronary artery semantic labeling.

Among graph matching-based methods, NGM, AGMN, and EAGMN perform vertex classification using association graph, while MGM conducts feature embedding and graph matching using the spectral method MatchEIG, preserving cycle consistency among multiple graphs. Results demonstrate that anatomical connections and cycle consistency between ICAs guide the matching process. The success of MGM reinforces the significance of learning coronary anatomy across different ICAs in clinical practice, where cardiologists acquire knowledge by comparing a test case to multiple reference cases in a template set. MGM outperformed the highest ACCs across all artery labeling types, highlighting the superiority of the usage of MGM for coronary artery semantic labeling.

### 4.5. Domain Difference Tests and Model Interpretation

**Uni-site test**. We conducted experiments to evaluate performance differences using ICAs from a single site. For each site, we used stratified sampling to divide the entire dataset into a training set and a template set, comprising 80% and 20% of the ICAs, respectively. The reason to increase the number of ICAs in the template set is due to the limited number of ICAs in the uni-site test. Within the training set, we further applied 5-fold cross-validation. The performance results are presented in Table 6.

**Table 6.** Achieved performance for coronary artery semantic labeling using ICAs from individual sites.

| Site | Artery type | LMA | LAD | LCX | D | OM | Avg |
|---|---|---|---|---|---|---|---|
| S1 | ACC | 0.9831±0.0138 | 0.9469±0.0183 | 0.9445±0.0264 | 0.9372±0.0285 | 0.8569±0.0385 | 0.9321±0.0182 |
|  | PREC | 0.9831±0.0138 | 0.9534±0.0217 | 0.9069±0.0256 | 0.9274±0.0207 | 0.9177±0.0299 | 0.9377±0.0175 |
|  | REC | 0.9831±0.0138 | 0.9469±0.0183 | 0.9445±0.0264 | 0.9372±0.0285 | 0.8569±0.0385 | 0.9337±0.0179 |
|  | F1 | 0.9831±0.0138 | 0.9501±0.0194 | 0.9249±0.0161 | 0.9323±0.0243 | 0.8856±0.0244 | 0.9352±0.0176 |
| S2 | ACC | 0.9736±0.0143 | 0.9730±0.0169 | 0.9620±0.0198 | 0.9310±0.0283 | 0.9661±0.0114 | 0.9736±0.0143 |
|  | PREC | 0.9734±0.0148 | 0.9597±0.0143 | 0.9557±0.0248 | 0.9597±0.0239 | 0.9685±0.0119 | 0.9734±0.0148 |
|  | REC | 0.9736±0.0143 | 0.9730±0.0169 | 0.9620±0.0198 | 0.9310±0.0283 | 0.9667±0.0122 | 0.9736±0.0143 |
|  | F1 | 0.9735±0.0142 | 0.9663±0.0150 | 0.9588±0.0221 | 0.9450±0.0246 | 0.9675±0.0121 | 0.9735±0.0142 |

According to Table 6, the performance for coronary artery semantic labeling was superior at Site 2, indicating that the model's effectiveness can vary depending on the site, possibly due to differences in data quality, imaging techniques, or patient demographics. However, the high performance metrics across both sites affirm the robustness of the proposed MGM for coronary artery semantic labeling.

**Cross-site test**. We conducted supplementary experiments with ICAs from either of the two hospitals separately. Additionally, to assess domain differences, we performed experiments where ICAs from one hospital were used for training and as template sets. Using the trained model, we tested the performance using ICAs from the other hospital. These groups of experiments evaluated the generalizability of the designed algorithms. The performance results are presented in Table 7.

**Table 7.** Achieved performance for coronary artery semantic labeling using cross-site ICAs for testing. The Tr&Tp indicates that the ICAs from Site 1 were used as the training and template sets, while Te indicates the site of ICAs from test set.

| Tr&Tp | Te | Artery type | LMA | LAD | LCX | D | OM | Avg |
|---|---|---|---|---|---|---|---|---|
| S1 | S2 | ACC | 0.9750±0.0172 | 0.9297±0.0128 | 0.9073±0.0158 | 0.9037±0.0164 | 0.8382±0.0203 | 0.9078±0.0101 |
|  |  | PREC | 0.9788±0.0175 | 0.9368±0.0131 | 0.8947±0.0137 | 0.8952±0.0212 | 0.8537±0.0215 | 0.9118±0.0082 |
|  |  | REC | 0.9750±0.0172 | 0.9297±0.0128 | 0.9073±0.0158 | 0.9037±0.0164 | 0.8382±0.0203 | 0.9108±0.0077 |
|  |  | F1 | 0.9769±0.0172 | 0.9332±0.0126 | 0.9009±0.0141 | 0.8994±0.0186 | 0.8459±0.0205 | 0.9113±0.0079 |
| S2 | S1 | ACC | 0.9843±0.0117 | 0.9136±0.0047 | 0.9078±0.0082 | 0.9029±0.0147 | 0.8338±0.0200 | 0.9025±0.0041 |
|  |  | PREC | 0.9843±0.0117 | 0.9276±0.0086 | 0.8923±0.0116 | 0.8740±0.0048 | 0.8662±0.0116 | 0.9089±0.0047 |
|  |  | REC | 0.9843±0.0117 | 0.9136±0.0047 | 0.9078±0.0082 | 0.9029±0.0147 | 0.8338±0.0200 | 0.9085±0.0046 |
|  |  | F1 | 0.9843±0.0117 | 0.9205±0.0046 | 0.8999±0.0079 | 0.8882±0.0077 | 0.8496±0.0134 | 0.9085±0.0047 |

According to Table 7, when training on ICAs from one site and testing on ICAs from another, the model achieved consistently high performance, with an ACC greater than 0.9. Specifically, using the training set from Site 1 and testing on ICAs from Site 2 resulted in an average ACC of 0.9078±0.0101. Conversely, training on Site 2 and testing on Site 1 produced an average ACC of 0.9025±0.0041. These results indicate that the model generalizes well across different domains, maintaining high performance despite variations between the two sites.

**Cross-site template**. We further conducted experiments using templates from another hospital to validate the influence of the selected template ICAs on model performance. According to Table 8, using cross-site template ICAs, the proposed MGM achieved consistently high performance, with an ACC of 0.9464 and 0.9238 on these two groups of experiments, respectively. By comparing Tables 7 and 8, it is evident that performance relies more on the source of the training and testing sets, while changing the source of templates has a minor influence on performance. Table 7 also validated that our model is robust on template selection.

**Table 8.** Achieved performance for coronary artery semantic labeling using cross-site ICAs as the template. The Tr&Te indicates that the ICAs from Site 1 were used as the training and template sets, while Tp indicates the site of ICAs from test set.

| Tr&Te | Tp | Artery type | LMA | LAD | LCX | D | OM | Avg |
|---|---|---|---|---|---|---|---|---|
| S1 | S2 | ACC | 0.9752±0.0267 | 0.9723±0.0151 | 0.9692±0.0146 | 0.9709±0.0180 | 0.8323±0.0766 | 0.9464±0.0172 |
| | | PREC | 0.9752±0.0267 | 0.9793±0.0156 | 0.9020±0.0436 | 0.9565±0.0257 | 0.9502±0.0139 | 0.9526±0.0136 |
| | | REC | 0.9752±0.0267 | 0.9723±0.0151 | 0.9692±0.0146 | 0.9709±0.0180 | 0.8323±0.0766 | 0.9440±0.0185 |
| | | F1 | 0.9752±0.0267 | 0.9757±0.0136 | 0.9335±0.0185 | 0.9635±0.0205 | 0.8850±0.0403 | 0.9466±0.0170 |
| S2 | S1 | ACC | 1.0000±0.0000 | 0.8976±0.0270 | 0.9438±0.0364 | 0.9419±0.0361 | 0.8757±0.0769 | 0.9238±0.0156 |
| | | PREC | 1.0000±0.0000 | 0.9590±0.0248 | 0.9209±0.0475 | 0.8596±0.0343 | 0.9099±0.0608 | 0.9299±0.0147 |
| | | REC | 1.0000±0.0000 | 0.8976±0.0270 | 0.9438±0.0364 | 0.9419±0.0361 | 0.8757±0.0769 | 0.9318±0.0151 |
| | | F1 | 1.0000±0.0000 | 0.9271±0.0231 | 0.9321±0.0420 | 0.8986±0.0319 | 0.8923±0.0690 | 0.9300±0.0150 |

Based on these three experiments, we conclude that the proposed MGM is robust, and the selection of templates and data domains has limited influence on model performance, with all experimental results showing an average accuracy (ACC) above 0.9. We hypothesize that it is because the arterial topology and anatomy is site independent, and they are essential for accurate coronary artery semantic labeling. To verify the significance of the approach and their values in graph matching, we employ our modified perturbation-based method, ZORRO [9,10,33], to explain the feature importance of the proposed MGM. ZORRO is designed to iteratively and recursively enhance graph representation learning by adding key features and nodes, based on a fidelity score that evaluates the difference between the original and updated predictions after masking out significant elements. The algorithm employs a hard mask, where a value of 0 indicates a non-selected feature and a value of 1 indicates a selected feature.

To adapt ZORRO for our needs, originally intended to explain GNNs for node classification, we modified the algorithm to clarify feature importance in our MGM context. Given that we transformed the graph matching problem into a vertex classification task using a set of graphs, we use a unified feature mask to simultaneously mask selected features for all ICA-derived graphs in $\mathbb{G}$. The retained features from these individual graphs are then used to perform graph matching. If removing this feature causes a significant drop in performance, it indicates that this is an important feature. We choose 600 graph matching set and derive the feature importance as shown in Figure 4.

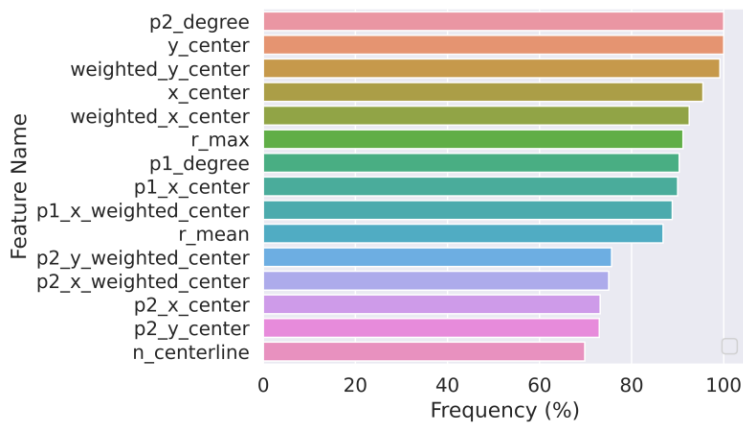

**Figure 4.** Feature importance ranking for classifying coronary arterial segments using MGM. Important features were surrogated by the frequency of the selected features during explaining all graph-matching pairs using the post-hoc GNN explaining algorithm, ZORRO. The vertical axis indicates the feature names, while the horizontal axis indicates the frequencies of the features, which are also denoted as feature importance.

In Figure 4, the frequency with which features are selected is used to denote their importance. For instance, the feature "p2_degree" had a selection frequency of 100%, indicating that it should be included every time MGM is performed, underscoring its significance. Among the top 15 features with the highest frequency and informativeness, all belong to the categories of topological and positional features. Specifically, "p2_degree" and "p1_degree" represent the degree of arterial segments, where the degree indicates the number of connections a segment has within the vascular tree-generated ICA graph. The "y_center" and "x_center," including weighted centers, indicate the position and coordinates of the arterial segments, while the p1 and p2 related centers refer to the position and coordinates of the endpoints of each arterial segment.

Figure 4 supports our hypothesis that the significance of topology in arterial identification motivates the transformation of arteries and their connections into graph structures. Consequently, integrating both topological and positional features for coronary artery semantic labeling is a practical and justified approach. This also explains why the uni-site test, cross-site template experiments, and cross-site test experiments achieved strong performance: despite significant variations in image quality across different institutions, the anatomy and template remained relatively consistent. Thus, incorporating topology and graph matching for semantic labeling is a reliable method for identifying individual coronary arteries.

### 4.6. Stenosis detection

We applied the proposed stenosis detection algorithm in Section 3.5 on the tested ICAs among the 5-fold cross-validation. We measured the stenosis detection performance using $ACC_s$ defined in Eq. 13, as shown in Table 9. We also visualized the stenosis detection results in Figure 4.

**Table 9**. Stenosis detection to identify all, minimal, mild, moderate and severe stenosis in terms of $ACC_s$.

| Artery Category | $ACC_s$ | # segments | Stenosis Type | $ACC_s$ | # segments |
|---|---|---|---|---|---|
| LMA | 0.9743 | 39 | Minimal (0% - 24%) | 0.8000 | 35 |
| LAD | 0.9492 | 256 | Mild (25% - 49%) | 0.8778 | 319 |
| LCX | 0.9094 | 243 | moderate (50% - 69%) | 0.9312 | 349 |
| D | 0.9315 | 190 | Severe (70%-100%) | 0.9666 | 209 |
| OM | 0.8478 | 184 | Overall | 0.9155 | 912 |
| Total | 0.9155 | 912 | | | |

The performance of the model in labeling coronary artery segments is evaluated by both artery category and stenosis type. The model achieves the highest accuracy ($ACC_s$) for the LMA at 0.9743, despite having the fewest segments (39). For the LAD artery, the accuracy is 0.9492 across 256 segments. The LCX and diagonal branches (D) show accuracies of 0.9094 and 0.9315, respectively, with a moderate number of segments. The OM artery has the lowest accuracy at 0.8478 with 184 segments. Overall, the model achieves an accuracy of 0.9155 across 912 segments. When categorized by stenosis type, the model performs best on severe stenosis (70%-100%) with an accuracy of 0.9666 for 209 segments, followed by moderate stenosis (50%-69%) at 0.9312 for 349 segments, mild stenosis (25%-49%) at 0.8778 for 319 segments, and minimal stenosis (0%-24%) at 0.8000 for 35 segments, demonstrating robust performance across varying degrees of arterial blockage.

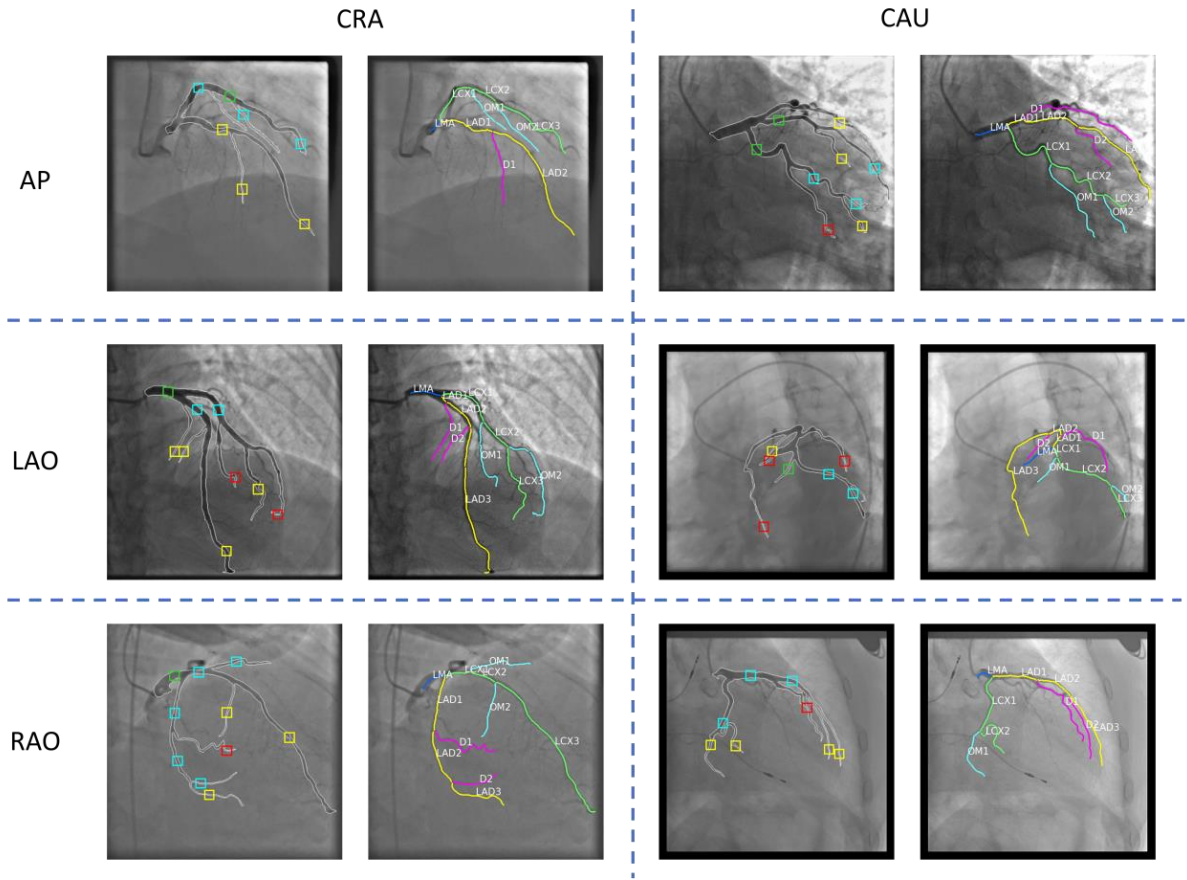

**Figure 4**. Stenosis detection results for representative ICAs under different view angles. The coronary artery contours and detected stenotic lesions are annotated in the left figure, while the semantic centerlines are shown in the right figure. The green, cyan, yellow, and red rectangles represent minimal, mild, moderate, and severe stenosis, respectively.

### 4.7. Clinical Application and Future work

ICA involves the injection of contrast media into the epicardial arteries with the acquisition of continuous fluoroscopy. Automatic identification of correct anatomical branches provides meaningful information for automated diagnosis, report generation and region of interest visualization [14]. Additionally, correct anatomic identification of severely diseased coronary arteries influences choice of treatment, for example whether or not a patient is a candidate for bypass surgery versus percutaneous coronary intervention. Proper labeling of main vessel and branch segments is crucial for patient treatment pathways. Successfully detecting the percent stenosis of a coronary artery branch improves diagnostic efficiency and confidence [34]. Thus, identifying individual coronary arterial segments from the vascular tree is important.

Physicians often base their assessment of blockages on the presence and severity of lesions within these main branches [35]. For example, to differentiate between various types of complex lesions, such as bifurcation, calcified, chronic total occlusions, and unprotected left main coronary artery lesions, it is required to extract LMA from ICA first and provide the extracted arterial segment for cardiologists for screening. Thus, semantic segmentation of the coronary arterial tree to extract individual coronary arterial branches is important.

Future work will focus on integrating the proposed semantic labeling techniques with wire-derived fractional flow reserve measurements to assess the functional significance of coronary lesions and guide revascularization decisions. Enhancing the algorithms to handle complex lesions and conducting extensive clinical validation will be essential. Additionally, exploring real-time processing, and personalized treatment planning will further improve the accuracy and utility of the methods. Emphasizing explainable AI models will ensure clinical trust and adoption, ultimately advancing coronary artery disease diagnosis and management.

## 5. Conclusion

In this paper, we present a multi-graph graph matching-based method for coronary artery semantic labeling. Intra-graph and cross-graph GCNs were adopted to perform feature embedding, and a spectrum-based method was employed to execute multi-graph graph matching while considering cycle consistency. We further evaluated the performance of stenosis detection using the semantic labeled arteries. This method not only opens new avenues for improved analysis of vascular anatomy but also lays the foundation for multi-ICA joint analysis. In the future, further exploration of coronary artery semantic understanding and functional measurement within ICA videos will be pursued.

**Acknowledgement**

This research was supported by a research seed fund from Michigan Technological University Health Research Institute, three NIH grants (1R15HL172198, 1R15HL173852, and U19AG055373) and an Interdisciplinary Seed Grants from Kennesaw State University (000149).

**Declaration of Competing Interest**

The authors declare that they have no known competing financial interests or personal relationships that could have appeared to influence the work reported in this paper.

**Credit authorship contribution statement**

Chen Zhao: Conceptualization, methodology, coding, manuscript writing.

Zhihui Xu: Data management and clinical validation.

Pukar Baral: Data management.

Michele Esposito: Clinical validation and manuscript writing.

Weihua Zhou: Supervision, project administration, funding acquisition, manuscript writing, and review.